\documentclass[lettersize,journal]{IEEEtran}
\usepackage{amsmath,amsfonts}
\usepackage{algorithmic}
\usepackage{algorithm}
\usepackage{array}
\usepackage{textcmds}
\usepackage{enumerate}
\usepackage{romannum}
\usepackage{hyperref}
\usepackage{tabularx}
\usepackage{booktabs} 
\usepackage{textcomp}
\usepackage{url}
\usepackage{verbatim}
\usepackage{graphicx}
\usepackage{cite}
\usepackage{cleveref}
\usepackage{hyperref}
\usepackage{caption} 
\usepackage{subcaption} 
\usepackage[utf8]{inputenc} 
\usepackage{amsmath} 
\usepackage{textcomp} 
\usepackage{balance}
\DeclareUnicodeCharacter{03B2}{$\beta$}
\balance
\begin{document}

\title{MMCFND: Multimodal Multilingual Caption-aware Fake News
Detection for Low-resource Indic Languages}


\author{
    \IEEEauthorblockN{Shubhi Bansal\IEEEauthorrefmark{1}, Nishit Sushil Singh\IEEEauthorrefmark{1}, Shahid Shafi Dar\IEEEauthorrefmark{1}, Nagendra Kumar\IEEEauthorrefmark{1}}
   \\ 
    \IEEEauthorblockA{\IEEEauthorrefmark{1}Indian Institute of Technology Indore
    \\\{phd2001201007, cse200001056, phd2201201004, nagendra\}@iiti.ac.in}
}




\maketitle
\thispagestyle{empty}
\begin{abstract}
The widespread dissemination of false information through manipulative tactics that combine deceptive text and images threatens the integrity of reliable sources of information. While there has been research on detecting fake news in high-resource languages using multimodal approaches, methods for low-resource Indic languages primarily rely on textual analysis. This difference highlights the need for robust methods that specifically address multimodal fake news in Indic languages, where the lack of extensive datasets and tools presents a significant obstacle to progress. To this end, we introduce the \textbf{M}ultimodal \textbf{M}ultilingual dataset for \textbf{I}ndic \textbf{F}ake \textbf{N}ews \textbf{D}etection (MMIFND). This meticulously curated dataset consists of 28,085 instances distributed across Hindi, Bengali, Marathi, Malayalam, Tamil, Gujarati and Punjabi. We further propose the \textbf{M}ultimodal \textbf{M}ultilingual \textbf{C}aption-aware framework for \textbf{F}ake \textbf{N}ews \textbf{D}etection (MMCFND). MMCFND utilizes pre-trained unimodal encoders and pairwise encoders from a foundational model that aligns vision and language, allowing for extracting deep representations from visual and textual components of news articles. The multimodal fusion encoder in the foundational model integrates text and image representations derived from its pairwise encoders to generate a comprehensive crossmodal representation. Furthermore, we generate descriptive image captions that provide additional context to detect inconsistencies and manipulations. The retrieved features are then fused and fed into a classifier to determine the authenticity of news articles. The curated dataset can potentially accelerate research and development in low-resource environments significantly. Thorough experimentation on MMIFND demonstrates that our proposed framework outperforms established methods for extracting relevant fake news detection features.
\end{abstract}
\section{Introduction}
\IEEEPARstart{T}{HE} proliferation of online fake news is a significant global challenge in today's digital landscape. These narratives exploit various formats, such as texts, images, and videos, to manipulate public opinion. They appear as fabricated stories, deceptive headlines, out-of-context information, and misconstrued satire. While this issue is prevalent worldwide, it poses a particularly acute threat in regions like India, which has diverse languages and high mobile phone penetration. India ranks among the countries with the highest levels of concern about online misinformation\footnote{https://www.statista.com/chart/31605/}. In the age of the internet, rumors, manipulated images, click-bait, biased stories, unverified information, and planted stories easily spread among India's 35 crore internet users. Social media platforms, particularly WhatsApp (29.8\%), Instagram (17.8\%), and Facebook (15.8\%)\footnote{http://timesofindia.indiatimes.com/articleshow/106899024.cms}, messaging applications, websites, and blogs disguising themselves as legitimate news sources become primary channels for disseminating fabricated information. Nearly 54\% of Indians rely on social media for news\footnote{https://www.thequint.com/tech-and-auto/tech-news/indians-misinformation-facts-social-media-oxford-university-press-study}, making them prime targets for fabricated content. Factors such as the private nature of messaging apps, algorithmic emphasis on engagement within social media algorithms, emotional resonance, confirmation bias \cite{nickerson1998confirmation}, and the isolating nature of social media echo chambers \cite{jamieson2008echo} contribute to the rampant spread of fake news. 

The unchecked spread of multimodal fake news has severe consequences for Indic languages. Communities that speak these languages often lack robust fact-checking resources, making them more susceptible to manipulation. Fake news can fuel existing social divides, potentially leading to unrest or violence. As manipulation techniques become more sophisticated, public trust in news and media declines, hindering a healthy democracy. Since trust and accurate information are crucial for economic and social progress, societies plagued by fake news struggle to attract investment and address complex issues such as poverty, public health, and education. Manual detection of fake news is labor-intensive, time-consuming, and prone to bias. Humans find it challenging to keep up with the constant influx of online content, especially across diverse languages. Analyzing large amounts of textual data and identifying subtle linguistic cues associated with deception is an enormous task. These factors underscore the critical need for robust automated fake news detection systems in Indic languages. Automated systems can identify potential fake news early, allowing for timely interventions and fact-checking. This proactive approach can help mitigate the spread of misinformation and foster a healthier online information environment.

\begin{figure*}
\subfloat[Politician Image with Inaccurate Information]{%
  \includegraphics[width=0.45\textwidth, height=6.5cm]{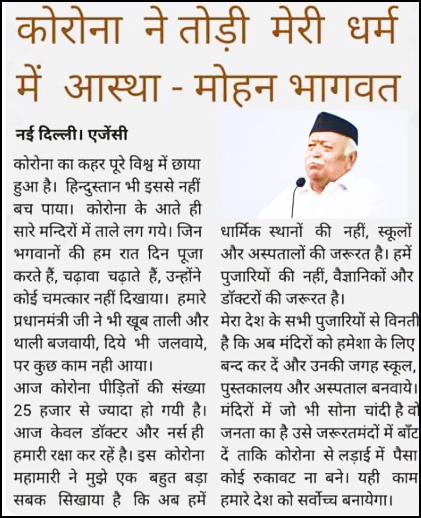}\label{fig:f1}%
}
\subfloat[Video Distorts Reason for Parrot Astrologer's Arrest]{%
  \includegraphics[width=0.45\textwidth,height=6.5cm]{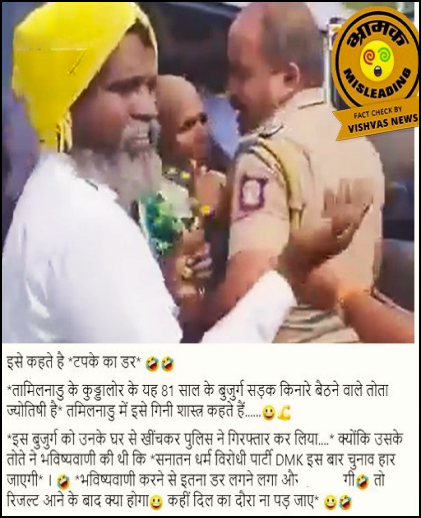}\label{fig:f2}%
}
\caption{Examples of Multimodal Fake News on Web Platforms in Indic Languages}
\label{fig:intro}
\end{figure*}
Prior methods of detecting fake news have primarily focused on multimodal approaches for high-resource languages \cite{zhang2023multimodal,sormeily2024mefand} and solely text-based approaches for low-resource languages \cite{thaokar2022multi,sivanaiah2022fake,warjri2023fake}. However, the spread of multimodal fake news is particularly dangerous in India, where there is linguistic diversity and a lack of reliable fact-checking resources for low-resource languages. Images have the ability to tap into our subconscious and enhance the perceived credibility of information, even when the accompanying text is questionable. This makes multimodal manipulation highly effective, as creators can use visuals to evoke strong emotional responses that bypass critical thinking, which is especially concerning in societies grappling with underlying tensions. Relying solely on text analysis poses its own challenges because a single image can completely distort the meaning of a statement or provide deceptive proof for a claim. Additionally, image editing techniques are becoming increasingly sophisticated, making it difficult to spot subtle alterations. Furthermore, online content in Indic languages often incorporates code-mixing and transliteration, further complicating text-based detection techniques. As shown in the example post (\autoref{fig:f1}), a viral image\footnote{https://www.vishvasnews.com/english/politics/fact-check-rss-chief-mohan-bhagwat-did-not-say-he-has-lost-faith-in-religion-due-to-corona/} circulated on Internet falsely claimed that Mohan Bhagwat, the leader of the Rashtriya Swayamsevak Sangh (RSS), expressed a loss of faith in religion due to the COVID-19 pandemic. The image was manipulated to resemble a newspaper clipping, adding credibility to the fabricated quote. This case highlights the significance of multimodal fake news detection as it combines a deceptive image with fabricated text to gain credibility. It also demonstrates how such manipulations can target specific audiences, such as Indic language communities where religion holds cultural importance. Another example (\autoref{fig:f2}) involves a video\footnote{https://www.vishvasnews.com/politics/fact-check-tamilnadu-cuddalore-fortune-teller-news-viral-with-misleading-claim/} circulating on social media, allegedly showing police arresting a parrot astrologer for his political predictions. However, this is fake news, as the arrest was related to illegally possessing a protected parrot. 
The video appears authentic and depicts a police arrest, but the text twists the narrative into a politically charged scenario. It requires careful multimodal analysis, considering both the visuals and the text within its cultural context, to uncover that the arrest was connected to animal protection laws, not political interference. These examples underscore how quickly false narratives can gain traction online, underscoring the importance of robust multimodal fake news detection tools for Indic languages. Consequently, it is crucial to develop multimodal fake news detection capabilities specifically for low-resource Indic languages. This approach would address the challenges posed by limited data availability and the unique complexities of these languages. By analyzing textual content alongside associated visuals, this research aims to enhance the accuracy of fake news detection in a context often overlooked by existing technologies.

Existing datasets for Indic languages tend to focus either on multilingual \cite{Praseed2023, sharma2023ifnd} or multimodal \cite{nakamura2019r,suryavardan2023factify} aspects, but not both. It creates a significant gap in understanding and combating fake news within these language communities. Fake news often leverages the power of images along with text to deceive and manipulate. Hence, a dataset that includes images and texts across seven low-resource Indic languages is crucial for studying this complex interplay of misinformation. Moreover, low-resource languages often lack advanced Natural Language Processing (NLP) tools and have limited training data, which makes the inclusion of images in a dataset even more vital. The consequences of not having specialized tools to understand and combat multimodal misinformation lead to the erosion of trust in institutions, fuel social unrest and exploit communities with limited media literacy. Therefore, a multimodal dataset is essential to develop effective detection models for Indic languages. Such a dataset will empower researchers to protect the low-resource language communities from the devastating effects of fake news.

Prior works for detecting fake news\cite{Purwanto2021, Zhou2023, Tahmasebi2023, Jiang2023} focus on contrastive learning, which examines the cosine similarity between visual and textual representations. Although these approaches emphasize the alignment between the two modalities, they cannot analyze how the image and text interact to form potentially inaccurate narratives. On the other hand, utilizing textual and visual features directly extracted from a foundation model with contrastive and multimodal capabilities offers distinct advantages for detecting multimodal fake news. This model allows for direct reasoning about the relationship between the image and text, enabling the detection of contradictions, subtle manipulations, or the reinforcement of false narratives. Additionally, this model facilitates cross-attention, helping it learn which specific parts of the image and text are most relevant to each other in deciphering the manipulative intent of multimodal fake news.

Caption generation is crucial in a multimodal approach to Indic fake news detection as it allows algorithms to analyze images and text together, addressing the limitations of text-only models. Caption generation provides insight into the manipulative strategies employed in Indic fake news. Fake news creators carefully construct the relationship between images and text to maximize impact. The generated caption serves as a bridge, revealing how the image is intended to support or distort the textual narrative. Since images can be inherently ambiguous and open to multiple interpretations, the generated caption acts as a guide, shaping the interpretation of potentially ambiguous visual content and reinforcing the distorted narrative. Inconsistencies or contradictions identified by comparing the generated caption with the textual content often indicate manipulation, as fake news creators struggle to maintain coherence between the visual and textual elements. While existing work \cite{Meel2021} leverages generated captions by tokenizing them and calculating cosine similarity with textual content, our approach primarily focuses on the semantic overlap between the two modalities. In contrast, including the generated caption directly as an additional feature preserves the intended manipulative relationship between the image and its accompanying caption created by the fake news creator. This allows for a holistic analysis of the interaction between visual and textual elements. Deep learning and transfer learning models can learn more complex and subtle patterns that indicate fake news in the Indic language context by including the raw caption alongside visual and textual data.

In this paper, we curate the first extensive \textbf{M}ultimodal \textbf{M}ultilingual dataset for \textbf{I}ndic \textbf{F}ake \textbf{N}ews \textbf{D}etection (MMIFND), encompassing seven low-resource Indic languages and containing roughly 28,085 instances of real and fake news samples. Furthermore, we propose a \textbf{M}ultilingual \textbf{M}ultimodal \textbf{C}aption-aware framework for \textbf{F}ake \textbf{N}ews \textbf{D}etection (MMCFND). 
We derive visual and textual feature representations using unimodal pre-trained encoders and pairwise foundation model encoders. The foundation model employs a two-stage encoding process. First, individual modalities are processed by separate encoders. Then, the resultant embeddings are combined using a fusion encoder, yielding a crossmodal representation. We also generate descriptive image captions, enriching the visual representation alongside the original textual and visual content. MMCFND’s four-pathway architecture extracts features from image, text, caption, and image-text modalities and feeds them into a classifier to distinguish between real and fake news articles. Our contributions can be summarized as follows:
\begin{itemize}
\item{In a pioneering effort to combat misinformation in the Indian subcontinent, we release the first-ever large-scale, multilingual, and multimodal dataset in seven under-resourced Indic languages. The MMIFND dataset consists of 28,085 instances meticulously segmented across Hindi, Bengali, Marathi, Malayalam, Tamil, Gujarati, and Punjabi. We believe this unique dataset will serve as a cornerstone for future research on fake news in the Indian landscape.}
\item{We present MMCFND, a new framework for detecting fake news in low-resource Indic languages using multimodal and multilingual approaches. Our methodology incorporates foundational model-based learning of semantic representations to quantify crossmodal similarity between textual and visual elements. This allows for the informed mapping and fusion of multimodal features.}
\item{We employ a vision-language model with a two-stage pre-training strategy to generate descriptive image captions. These captions, along with textual and visual content, are strategically used in our framework to detect false news. Captions enable a deeper contextual understanding by providing a linguistic representation of visual elements, helping detection models identify potential contradictions or deliberate reinforcement between textual and image-based components.}
\item{Rigorous experimentation on two distinct datasets confirms that features retrieved from foundational model enhance unimodal features with crucial additional information. As a result, our proposed MMCFND framework outperforms existing fake news detection methods.}
\end{itemize}
The remainder of this paper is structured as follows. \Cref{sec:rw} reviews relevant research in the field. \Cref{sec:pd} offers a precise definition of the problem this work addresses. \Cref{sec:meth} details the proposed methodology.  Experimental evaluations are then presented in \Cref{sec:exp}. \Cref{sec:con} summarizes key insights and potential future research directions.
\section{Related Works}
\label{sec:rw}
In the pursuit of mitigating the proliferation of misinformation, contemporary research has witnessed a burgeoning interest in various facets of fake news detection. This section delves into the existing literature, categorized into three key areas: multilingual fake news detection, multimodal fake news detection, and fake news datasets specifically designed for the Indian context.
\subsection{Multilingual Fake News Detection}
Social media users post content in their native regional langauges to express opinions on trending topics \cite{dar2024social, dar2024contrastive, bansal2024multilingual, rehman2023user}.
Given the global nature of online communication, detecting false information in diverse languages is crucial. This study examines existing methodologies for identifying fake news in multilingual contexts. Raja \textit{et al}. \cite{Raja2023} used transfer learning with pre-trained multilingual models such as mBERT and XLM-R to classify fake news in Dravidian languages at the sentence level. The effectiveness of transfer learning was assessed by fine-tuning the pre-trained models on a combination of English and Dravidian fake news datasets. Tufchi \textit{et al.} \cite{tufchitransvae} proposed TransVAE-PAM, which used transformer models to create informative embeddings of news articles. The authors employed Variational Autoencoder (VAE) and its β-variant for dimensionality reduction and the Pachinko Allocation Model (PAM) to extract important topics from news articles. 
Sharma \textit{et al.} \cite{sharma2023lfwe} introduced the LFWE model for detecting Hindi fake news. Using 23 linguistic features (including 8 novel features), the authors created word embeddings categorized into six linguistic dimensions. The LFWE model, trained on these features, achieved the highest accuracy with a Support Vector Classifier on the newly curated HinFakeNews dataset. Bala \textit{et al.} \cite{bala2023abhipaw} fine-tuned MuRIL using supervised learning to classify comments and posts as either authentic or fake. The authors observed improved performance in detecting Dravidian fake news, validating the model's ability to learn linguistic nuances and contextual aspects of Dravidian languages. Mohawesh \textit{et al.} \cite{Mohawesh2023} introduced a novel multilingual capsule network model incorporating multilingual embeddings with semantic infusion. The architecture utilized parallel channels for source (English) and target languages (English, Hindi, Indonesian, Swahili, and Vietnamese). Contextual features are extracted through a Bidirectional LSTM (BiLSTM), while a capsule network models hierarchically organized dependencies and relationships. Mundra \textit{et al.} \cite{Mundra2023} proposed a two-phase machine intelligence framework for fake news detection. In the first phase, semantic features are extracted using MLP, CNN, and BiLSTM initialized with GloVe and BERT embeddings. In the second phase, a selection of the top four models from first phase is combined using various ensemble techniques (voting, bagging, boosting, and stacked ensembles) where the soft-weighted voting ensemble achieved the highest performance. 
Shailendra \textit{et al.} \cite{Shailendra2022} demonstrated the efficacy of a novel CNN and BiLSTM architecture in the domain of Hindi fake news detection. By leveraging Facebook's fastText embeddings, the proposed model achieved performance gains exceeding both a baseline model and a BiLSTM model employing the same embeddings.

While existing studies have addressed multiple languages, they often prioritize major languages or language families, potentially neglecting low-resource languages or specific linguistic communities.
By encompassing seven low-resource Indic languages in the MMIFND dataset and MMCFND framework, we ensure that Indic language speakers' linguistic diversity and needs are adequately represented. This inclusivity is essential for addressing the challenges of fake news detection in linguistically diverse societies.
\subsection{Multimodal Fake News Detection}
Multimodal data leverages a variety of formats such as combination of textual content, images, and videos \cite{bansal2022hybrid, rehman2025context, anshul2023multimodal, bansal2024hybrid, dar2024contrastive} to misguide audiences. To this end, we conducted research on multimodal fake news detection techniques that exploit these modalities. Meel \textit{et al.} \cite{Meel2021} developed a multimodal framework for fake news detection that combines textual and visual analysis. Their approach uses a Hierarchical Attention Network (HAN) for in-depth textual analysis, an image captioning tool to generate image summaries, and forgery analysis techniques to detect image manipulations. The final classification decision is made through a max-voting ensemble approach that integrates these diverse feature sets. Wu \textit{et al.} \cite{Wu2021} introduced Multimodal Co-Attention Networks (MCAN), a novel approach for multimodal feature fusion in fake news detection. The authors employed three distinct sub-networks for textual, spatial, and frequency -related features, respectively. These features were then integrated using stacked co-attention layers, the design of which mimics iterative processing and fusion of visual and textual information commonly observed when individuals evaluate news. Yadav \textit{et al.} \cite{Yadav2023} devised a transformer-based multilevel attention architecture that captures complementary information across modalities. Visual attention focuses on crucial image regions guided by attended textual features. Furthermore, self-attention is employed to eliminate redundancy within the fused multimodal data.
Zhou \textit{et al.} \cite{Zhou2023} presented FND-CLIP that leverages Contrastive Language-Image Pretraining (CLIP) for aligned multimodal feature extraction, guiding crossmodal learning. FND-CLIP extracts rich representations from news using unimodal encoders and pairwise CLIP encoders, weighting CLIP-generated features by inter-modality similarity. A modality-wise attention mechanism further enhances feature aggregation by reducing noise and redundancy in aggregated features. Jing \textit{et al.} \cite{Jing2023} proposed a Progressive Multimodal Fusion Network that captures textual and visual data at various levels. A mixer component enables fusion across these levels, strengthening inter-modality links. A transformer extracts visual features progressively, combining them with text features and image frequency information for fine-grained modeling. Finally, a novel fusion approach further enhances these connections. Hua \textit{et al.} \cite{Hua2023} introduced a novel BERT-based Back-Translation Text and Entire-image Multimodal model with Contrastive Learning (TTEC). TTEC employed back-translation on news text  to learn generalizable topic-related features. Next, a BERT-based model processed both textual and visual features, generating multimodal representations. Finally, contrastive learning is employed to refine these representations, leveraging similar news articles from the past. 

While existing frameworks utilize various machine learning, deep learning, transfer learning, and contrastive learning architectures, they often lack architectures specifically tailored to the unique challenges of multimodal fake news detection in the context of low-resource langauges. MMCFND addresses this gap with a four-pathway architecture designed to extract features from image, text, caption, and image-text modalities. The architecture's focus on multimodal and multilingual representations enhances classification accuracy and cross-language generalization. Unlike some multimodal approaches, MMCFND emphasizes the generation of descriptive image captions and the coherent integration of textual and visual modalities. The explicit focus on captions enriches the visual representation alongside visual and textual content, leading to improved accuracy and robustness in fake news detection.
\subsection{Fake News Datasets for Indian Context}
The nuances of language and cultural context can significantly impact fake news detection. In this section, we shed light on publicly available datasets for fake news detection within the Indian context. Raja \textit{et al.}\cite{Raja2023} contributed the Dravidian\_Fake dataset, a novel resource for classifying fake news in four Dravidian languages (Telugu, Tamil, Kannada, and Malayalam). To facilitate cross-lingual generalizability, the authors constructed multilingual datasets by merging Dravidian\_Fake with the existing English ISOT~\cite{ahmed2017detection} dataset. Addressing the scarcity of Hindi fake news resources, Badam \textit{et al.} \cite{Badam2022} compiled a valuable dataset containing 6,487 fake and 6,707 authentic news articles from AltNews that specializes in debunking misinformation spread by activists and media outlets. The authors employed various Machine Learning (ML) models on this dataset to determine their efficacy in Hindi fake news detection. Chakravarthi \textit{et al.}\cite{chakravarthi2024dataset} presented a pioneering dataset for Malayalam fake news categorization by distinguishing content based on its degree of falsehood. This dataset is the first of its kind for the Malayalam language. Tufchi \textit{et al.}\cite{Tufchi2023} introduced the FakeRealIndian Dataset (FRI) to facilitate research on fake news detection in the Indian context. FRI comprises 20,916 manually labeled articles (real and fake) sourced from various fact-checked platforms, including authentic news from the Times of India. FRI is the first publicly available Indian dataset encompassing both real and fake news narratives. The authors performed comprehensive preprocessing on the dataset and generated custom word embeddings. The richness of FRI makes it a valuable resource for researchers investigating the interplay of language, cultural nuances, and context in the realm of Indian fake news detection. 
Sharma\textit{ et al.} \cite{sharma2023lfwe} curated a novel annotated Hindi Fake News (HinFakeNews) dataset of roughly 33,300 articles. This dataset provides includes a rich array of information such as article text, topical categories, publication timestamps, authorship information, URLs, and additional metadata for analyzing Hindi fake news. Sharma \textit{et al.} \cite{sharma2023ifnd} introduced the Indian Fake News Dataset (IFND), a comprehensive resource for evaluating fake news detection models in India. Spanning 2013-2021 events, IFND includes texts, visuals, and even uses an algorithm to generate realistic fake news, enriching its representation. As the first large-scale Indian dataset of its kind, IFND provides valuable insights for researchers studying the complexities of Indian fake news. Singhal \textit{et al.}\cite{Singhal2022} introduced FactDrill, a novel and comprehensive data repository of fact-checked social media content specifically designed to investigate fake news phenomena in India. FactDrill leverages 22,435 samples curated from IFCN-affiliated Indian fact-checking websites, encompassing news stories from 2013 to 2020 in thirteen languages. The dataset encompasses  fourteen attributes, categorized into meta, textual, media, social, and event features. 
\cite{Dhawan2022} created a novel dataset, FakeNewsIndia, comprising 4,803 fake news incidents in the Indian context between June 2016 and December 2019. The authors established an automated data collection pipeline to harvest fake news incidents documented by Indian fact-checkers. De \textit{et al.} \cite{De2022} presented a multilingual and multi-domain dataset designed to address challenges in fake news detection for resource-scarce languages. The dataset encompasses English and four Asian languages (Hindi, Indonesian, Swahili, and Vietnamese), spanning seven distinct domains. To facilitate the inclusion of these under-resourced languages, the authors leverage Google Translate to render 480 news items from FakeNewsAMT and 500 from Celebrity Fake News. Rahman \textit{et al.} \cite{rahman2022fand} developed the Bengali Fake News Corpus (BFNC), a comprehensive annotated dataset designed for Bengali fake news detection. This corpus encompasses 5,178 news samples collected between January and August 2021.

While existing studies have contributed valuable datasets for fake news detection in multilingual contexts, they typically rely on datasets that are limited in scope or focus on specific language families or regions. By curating the MMIFND dataset, we address the need for a comprehensive dataset that spans seven low-resource Indic languages. This fills a crucial gap in the existing research landscape by providing a diverse and extensive dataset tailored specifically for multimodal fake news detection in low-resource Indic languages.
\section{Problem Definition}
\label{sec:pd}
This section provides a formal problem definition.
Let $A = \{a_i^{text}, a_i^{visual}: i=1,2,\hdots, N\}$ be the dataset containing a collection of multimodal news articles where $N$ represents the total number of articles and $L=\{l_1, l_2,\hdots, l_7\}$ be a set containing the seven low-resource Indic languages under consideration. Each element $a_i$ within the dataset $A$ is a tuple containing two components: $a_i^{text}$ that represents the textual content of the news article in a specific language $l \in L$ and $a_i^{visual}$ represents the visual information associated with the news article.
Mathematically, the problem can be expressed as:
\begin{equation}
f : A \rightarrow \{0, 1\}
\end{equation}
The core objective of this task is to develop a model, denoted as $f$, that can analyze a multimodal news article $a_i$ and predict its corresponding label $\hat{y}(a_i)$. We refer to the model's output as $\hat{y}(a_i)$ where $\hat{y}(a_i) \in \{0, 1\}$.
The function $\hat{y}: A \rightarrow \{0, 1\}$ works as follows:
$\hat{y}(a_i)$ = 0 signifies that news article $a_i$ is predicted as fake news, and $\hat{y}(a_i)$ = 1 signifies that news article $a_i$ is predicted as real news.
\section{Methodology}
\label{sec:meth}
\begin{figure*}
\includegraphics[width=\textwidth]{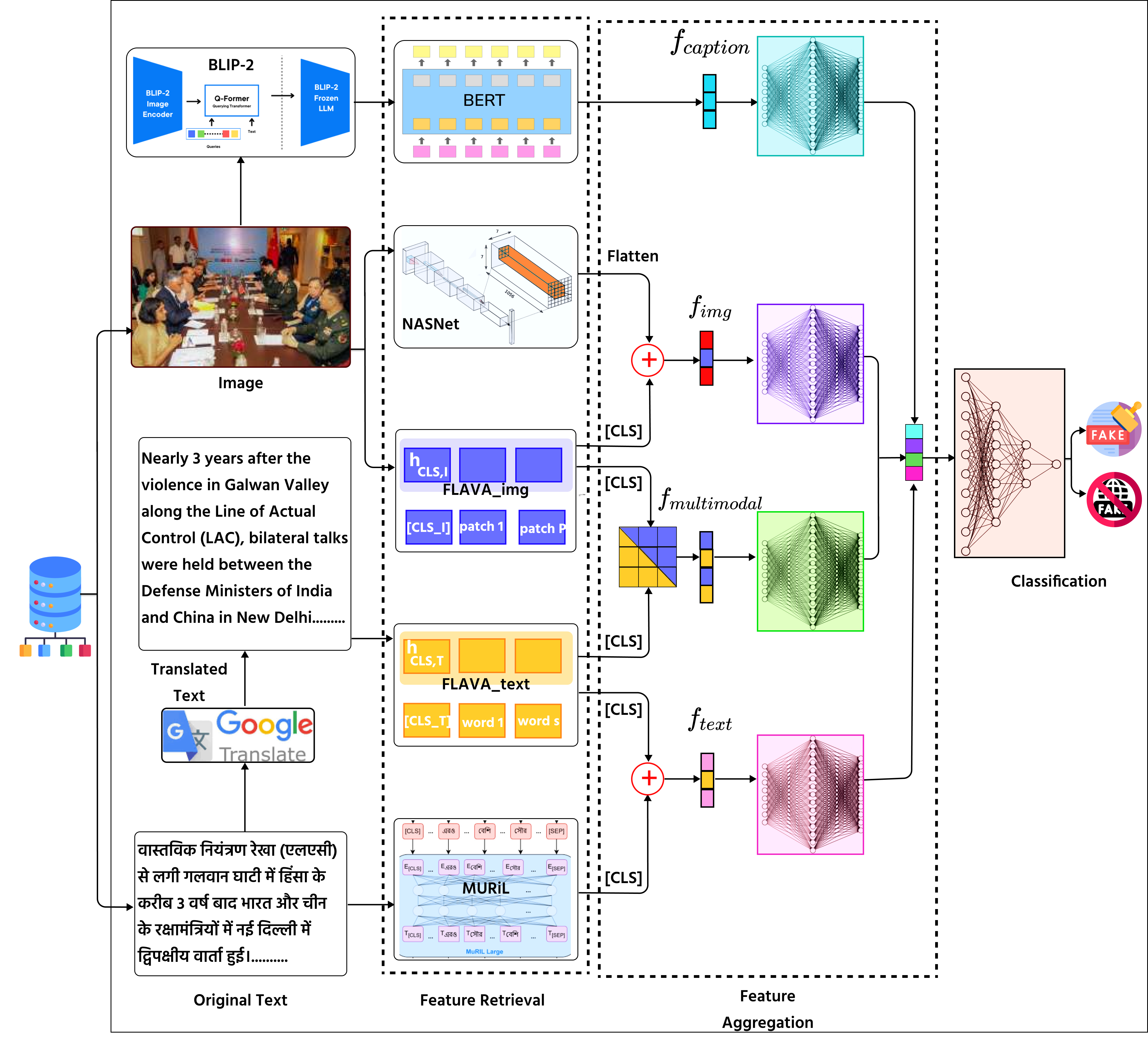}
\caption{System Overview}
\label{fig:overview}
\end{figure*}

In this section, we present our proposed methodology. The overall structure of our fake news detection system is outlined in \autoref{fig:overview}. The pipeline consists of three phases: feature retrieval, feature aggregation, and classification. In the feature retrieval module, features such as caption, image, text, and image-text are extracted from textual articles and associated images. The feature aggregation module uses fully connected layers to combine the outputs from the feature retrieval module. Finally, the integrated features are processed through the dense layers in the fake news detector to predict the authenticity of the content.
This section commences with a description of our data collection procedures, followed by a comprehensive analysis of the proposed model's architecture. 
\subsection{Data Curation}
\begin{figure*}[t]
\includegraphics[width=\textwidth]{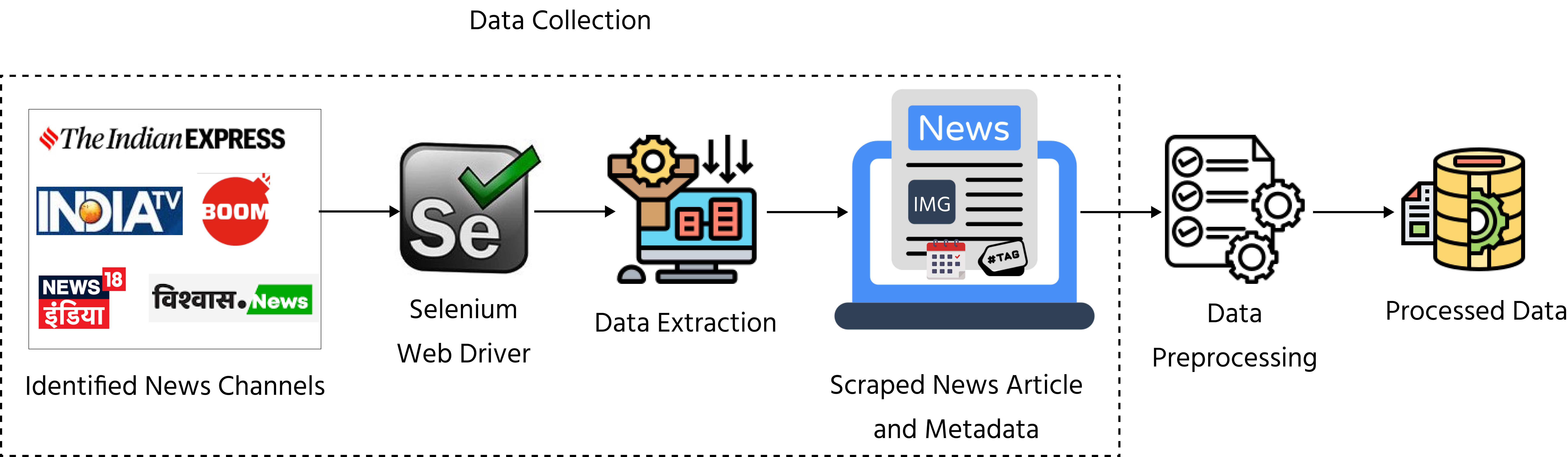}
\caption{Dataset Curation Pipeline}
\label{fig:dc}
\end{figure*}

This section discusses the data curation procedure, which consists of two stages. \autoref{fig:dc} illustrates the complete data curation process where the first stage focuses on data collection, followed by a dataset preprocessing stage.
\subsubsection{Data Collection}
To create a comprehensive dataset for detecting fake news in Indic languages, we used web scraping techniques to gather news articles from prominent Indian news channels. We prioritized sections of these channels that cover news from diverse categories to ensure a well-balanced representation of topics. In cases where a dedicated section for this content was unavailable, we extracted information from the national news section. Our selected sources included reputable platforms such as Boom\footnote{www.boomlive.in}, The Indian Express\footnote{www.indianexpress.com}, IndiaTV\footnote{www.indiatv.in}, News18\footnote{www.news18.com}, and Vishvas News\footnote{www.vishvasnews.com}.
\\

We chose these sources because they were easy to scrape and provided efficient data collection. However, we faced a significant challenge due to the prevalence of dynamic loading mechanisms and JavaScript-driven content presentation used by these websites. Traditional web scraping tools could not capture the complete content of these dynamic pages. To tackle this issue, we utilized Selenium\footnote{https://www.selenium.dev/}, a web automation framework known for its adaptability and effectiveness in handling the interactive nature of contemporary online news platforms. Our scraping process targeted various attributes such as textual content, images, publication dates, news tags, and keywords, which were extracted from the metadata of webpages. This meticulous approach resulted in the accumulation of 28,085 articles spanning seven Indic languages. For each dataset, we used the body or text as input and assigned a label (0 for fake and 1 for real) for classification.
\subsubsection{Dataset Preprocessing} 
To ensure consistency and cleanliness of the dataset, our data processing pipeline includes preprocessing procedures for raw textual and image data obtained from news websites. This allows for coherent analysis and extraction of relevant information. When scraping data from the web, raw data often contains artifacts such as Unicode characters, spaces, emojis, and URLs, which can introduce unnecessary noise. To enhance the reliability of subsequent analyses, we systematically remove these elements. The initial preprocessing stages for textual content involve removing Unicode characters to ensure consistency and avoid encoding discrepancies. We then clean the text by removing unnecessary spaces at the beginning and end and within it. This promotes standardized representations suitable for NLP tasks. Emojis and URLs are eliminated to prioritize textual purity and remove non-textual content. Additionally, we remove unnecessary spaces within the text to ensure homogeneity and coherence. Furthermore, articles without accompanying images are excluded from the dataset. Images are resized to dimensions (224, 224, 3) to ensure uniformity and compatibility with subsequent image processing tasks. 
Since the Foundational Language And Vision Alignment (FLAVA) model does not support Indian languages, we translated the original Indic language dataset into English using Google Translate\footnote{https://translate.google.co.in/} to align it with the linguistic capabilities of FLAVA. This dual approach resulted in a refined and standardized dataset of Indic and English textual components. This preprocessing strategy, covering both textual and image data, resulted in a refined and standardized dataset suitable for conducting insightful analyses of scraped news articles from various sources.
\begin{figure*}
\includegraphics[width=\textwidth,height=6cm]{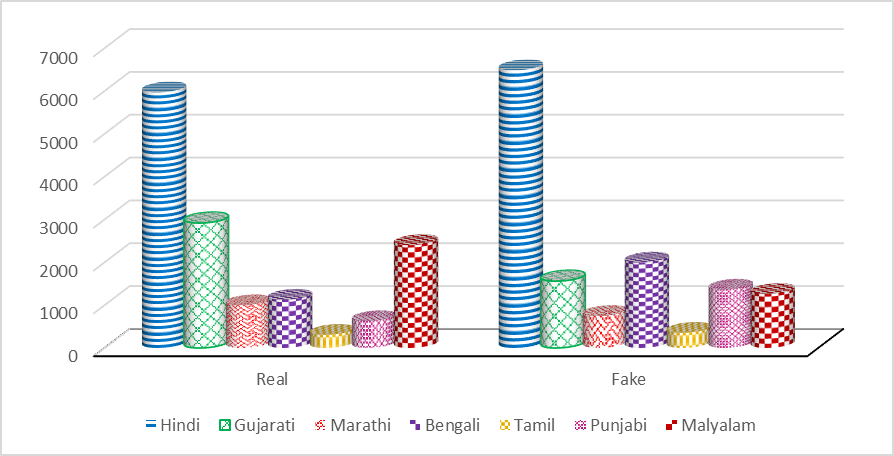}	\caption{\centering Language-wise Distribution of Real and Fake News Samples in MMIFND}
\label{fig:stats_mmifnd}
\end{figure*}
The distribution of the dataset among different languages is shown in \autoref{fig:stats_mmifnd}, demonstrating its compatibility with linguistic requirements and model specifications. Overall, this preprocessing approach provides a strong foundation for our research endeavors.

\subsection{Feature Retrieval}
In this section, we unfold the details of monomodal and FLAVA-guided feature retrieval.
\subsubsection{Monomodal Feature Retrieval}
This section elucidates procedures for retrieving features from individual modalities\\

(caption, text, and image) of the input multimodal news sample.
\paragraph{\textit{Feature Retrieval through Captions}} Social media platforms frequently utilize attention-grabbing titles and images to increase website traffic. However, in articles that are deliberately designed to deceive, the textual content often exhibits a significant discrepancy with the accompanying title and image. This deceptive practice is commonly termed ``clickbait".  Image captioning, the automated generation of descriptive text for an image is vital in combating clickbait. By providing users with an accurate summary of an image's content, image captioning empowers informed decision-making about whether to engage with the associated article. Our work leverages the Bootstrapping Language-Image Pre-training (BLIP-2) model~\cite{li2023blip}, a versatile visual-language model demonstrating exceptional performance across various multimodal tasks, including image captioning. BLIP-2's innovative pre-training paradigm involves generating synthetic captions and a filtering mechanism to refine them. It's ability to bridge visual and linguistic representations effectively makes it a compelling choice for generating comprehensive and informative captions from images.

BLIP-2's architecture comprises three primary modules: a vision encoder, a Querying Transformer (Q-Former), and a language model. The vision encoder extracts salient visual features from the input image, while the Q-Former is an intermediary between the visual and language modalities. This module transforms visual features into a representation suitable for processing by the language model, generating a coherent and semantically consistent caption. BLIP-2's caption generation process can be described as follows. First, the input image undergoes preprocessing (resizing and normalization) to conform to the model's requirements. Next, a meticulously designed image encoder extracts key visual features, creating a condensed representation of the image's visual content. This representation is then fed into the decoder module, which utilizes a transformer-based architecture with attention mechanisms to generate tokens corresponding to caption fragments. Finally, these tokens are decoded into natural language, forming a cohesive caption that accurately reflects the image's content. This initial caption generated by BLIP-2 is likely a raw textual description. The final features are obtained using the Bidirectional Encoder Representations from Transformers (BERT)~\cite{devlin2019bert} model as shown in \autoref{eq:caption}.
\begin{equation}
f_{caption} = BERT(BLIP-2(I)) 
\label{eq:caption}
\end{equation}
Here, $f_{caption} \in \mathbb{R}^{1 \times D}$, where $D$=768 denotes the embedding dimension of the generated caption. BERT has a deeper understanding of language nuances, allowing it to refine the caption's representation and create a richer and more meaningful feature vector.
\paragraph{\textit{Textual Feature Retrieval}}
Extracting salient features from textual content plays a pivotal role in fake news detection. Our research employs a two-pronged approach, meticulously tailored to address the multilingual nature of our dataset. For Indic text $(T_{Indic})$, we leverage the Multilingual Representations for Indian Languages (MuRIL) model~\cite{khanuja2021muril}, while English text $T_{English}$ is processed using the Foundational Language And Vision Alignment (FLAVA) model. This strategy ensures optimal feature representation by capitalizing on MuRIL's expertise in Indic languages and FLAVA's robust English language capabilities.

MuRIL, a BERT-based model, has been pre-trained on a corpus of 17 Indian languages. Its transformer architecture employs attention mechanisms to discern complex relationships between words,  enabling it to capture nuanced contextual information and generate highly informative word embeddings. FLAVA, a large-scale multimodal language model, demonstrates proficiency in processing images and text. Trained on a vast dataset of image-text pairs, FLAVA develops robust representations across both modalities, making it well-suited for analyzing English text in the context of fake news detection.

We adopt a consistent methodology to standardize the tokenization process for both MuRIL and FLAVA. A class token ([CLS]) is inserted at the beginning of the text, followed by a separator token ([SEP]) to delineate the sentence boundaries. This yields the pre-tokenized text, represented by the following equations:
\begin{equation}
    T_{Indic-tok}=MR\_Tok([CLS]+T_{Indic} +[SEP])  
\end{equation}
\begin{equation}
     T_{English-tok} = FL\_Tok( [CLS]+T_{English}+[SEP] ) 
\end{equation}
Here, $MR\-Tok$, $FL\_Tok$, $T_{Indic\_tok}$, and $T_{FLAVA\_tok}$ denote the MURIL and FLAVA tokenizers and their tokenized outputs, respectively. The special tokens ([CLS] and [SEP]) serve as clear markers for the models during processing. The tokenized Indic and English text ($T_{Indic-tok}$, $T_{English-tok}$) are then fed into their respective models (MuRIL and FLAVA) as shown below to extract the last hidden state representations.
\begin{equation}
     f_{MuRIL-text-lhs} = MuRIL( T_{Indic-tok})   
\end{equation}
\begin{equation}
    f_{FLAVA-text-lhs} = FLAVA\_Text( T_{English-tok} ) 
\end{equation}                            
Here, $f_{MuRIL-text-lhs} \in \mathbb{R}^{N\times D}$,  $f_{FLAVA-text-lhs} \in \mathbb{R}^{N \times D}$  represent the last hidden state outputs, with $N$ denoting the maximum number of tokens (200 in our case) and $D=768$ the dimensionality of each token's vector representation. The input text is either truncated or padded to ensure a uniform length. Finally, we obtain the classification token from $f_{MuRIL-text-lhs}$ and  $f_{FLAVA-text-lhs}$ giving $f_{MuRIL-text-cls} \in \mathbb{R}^{1 \times D}$,  $f_{FLAVA-text-cls} \in \mathbb{R}^{1 \times D}$.
The obtained representations are concatenated to create the final text embeddings as shown in \autoref{eq:fusion}.
\begin{equation}
     f_{text}=concat(f_{MuRIL-text-cls}, f_{FLAVA-text-cls)}  
\label{eq:fusion}
\end{equation}
Here, $f_{text}$ represents the overall textual feature vector of a multimodal news article. MuRIL's specialized knowledge of Indic languages and FLAVA's broader English understanding provide a robust analysis of the multilingual text in fake news. These models capture different aspects of language, leading to a more holistic textual feature representation.
\paragraph{\textit{Visual Feature Retrieval}}
To extract meaningful representations of images within multimodal news articles, we employ a dual-model approach featuring two state-of-the-art feature extraction models: FLAVA and \textit{N}eural \textit{A}rchitecure \textit{S}earch \textit{Net}work (NASNet). This strategy ensures robust visual analysis, capturing low-level and high-level image characteristics.

First, FLAVA preprocesses the input image by resizing and segmenting it into patches. These patches are then linearly embedded to obtain vector representations. Positional embeddings and a designated image classification token are appended to these vectors, serving as input to FLAVA's transformer-based encoder. We isolate the hidden state vector corresponding to the image classification token as exhibited in \autoref{eq:flava_image}.
\begin{equation}
f_{FLAVA-img} = (FLAVA\_Image(I))_{cls} 
\label{eq:flava_image}
\end{equation}
Here, $I$ represents the input image, and the subscript $cls$ indicates the classification token's output. Isolating FLAVA's image classification token ensures a condensed, high-level summary of the image's visual information. Simultaneously, we utilize NASNet~\cite{zoph2018learning} for its exceptional performance in image analysis. NASNet meticulously processes the input image through hierarchical layers, culminating in extracting low-level (edges, textures) and high-level (object recognition) visual features. We obtain the unimodal image representation from the penultimate layer of NASNet, referred to as the global average pooling layer, as shown in \autoref{eq:nasnetfe}.
\begin{equation}
    f_{NASNet-img} = NASNet(I)   
    \label{eq:nasnetfe}
\end{equation}
Here, $f_{NASNet-img}$ represents the unimodal image feature vector extracted from NASNet. Finally, we concatenate the feature vectors extracted by both models, yielding a comprehensive final image representation as depicted in \autoref{eq:imgfe}.
\begin{equation}
    f_{img} = concat( f_{NASNet-img}, f_{FLAVA-img}) 
    \label{eq:imgfe}
\end{equation}
Here, $f_{img}$ denotes the overall visual feature vector obtained by combining the two feature vectors, $f_{NASNet-img}$ and $f_{FLAVA-img}$, into a single vector. The concatenation of outputs provides a richer and more nuanced representation of the image content. FLAVA's transformer-based architecture excels in understanding relationships between image patches while NASNet effectively captures low-level details and hierarchical visual patterns. 
\subsubsection{FLAVA-guided Feature Retrieval}
The extraction of salient multimodal features hinges on the FLAVA model's multimodal encoder, which simultaneously processes both English text $(T_{English})$ and the corresponding image ($I$). FLAVA applies learned linear projections to each hidden state vector within the text and image representations to facilitate a seamless fusion of these modalities. These projections essentially refine the individual representations, preparing them for optimal integration. Crucially, a designated classification token ($[CLS_M]$) is incorporated into the unified list of projected vectors. This special token plays a vital role by enabling the model to execute cross-attentional computations. Cross-attention allows FLAVA to discern intricate relationships and dependencies between the projected text and image representations, fostering a comprehensive understanding of the multimodal input. Formally, we aim to extract the classification token from the FLAVA multimodal encoder's output, as exhibited in \autoref{eq:flavacmfe}.
\begin{equation}
f_{multimodal} = (FLAVA\_Multimodal(T_{English}, I ))_{cls}
\label{eq:flavacmfe}
\end{equation}
Here, $f_{multimodal}$ denotes the FLAVA-guided multimodal feature vector, encompassing the essence of textual and visual information within the news article. The subscript $`cls'$ emphasizes the output associated with the classification token, which serves as a condensed representation summarizing the core meaning derived from combined modalities. This fusion process, empowered by learned linear projections and cross-attention with the $[CLS_M]$ token, allows the model to exploit complementary information from text and image, potentially leading to a more robust and nuanced feature representation for fake news detection.
\subsubsection{Feature Aggregation}
In this stage, we strategically combine the diverse feature representations extracted in previous steps. Text features ($f_{text}$), derived from MuRIL and FLAVA\_Text, image features ($f_{img}$) obtained from NASNet and FLAVA\_Image, multimodal features ($f_{multimodal}$) generated by FLAVA\_Multimodal, and caption features ($f_{caption}$) produced by BLIP-2 are meticulously integrated. To optimize compatibility, each feature set is processed through individual fully connected layers. These layers refine the individual representations before concatenation. The resulting refined representations are then concatenated to obtain a comprehensive, unified feature vector as demonstrated in \autoref{eq:fagg}.
\begin{equation}
    f_{combined} = concat( f_{text}, f_{img}, f_{multimodal}, f_{caption})
    \label{eq:fagg}
\end{equation}                      
The concatenated feature vector ($f_{combined}$) embodies a rich fusion of salient characteristics from each modality, offering a nuanced and multifaceted representation of the multimodal news article. Concatenating diverse feature sets leverages the complementary information captured by each modality. This integrated representation offers a more holistic understanding of the news article compared to relying on individual modalities alone. To further refine this combined representation, enhance its discriminative power, and prepare it for final classification, we employ additional fully connected layers. These layers are instrumental in identifying complex patterns and dependencies within the fused features. Finally, the obtained output is fed to a classifier for distinguishing between real and fake news instances.
\subsubsection{Classification}
We feed the refined representation into a classifier to predict the label $\hat{y}$. To this end, we employ the binary cross-entropy loss ($L_{CE}$) as our objective function, a widely established metric for classification tasks.  Formally, the binary cross-entropy loss is defined as:
\begin{equation}
L_{CE} = y \log (\hat{y}) + (1 - y) \log (1 - \hat{y})
\end{equation}
Here, $L_{CE}$ is the binary cross-entropy loss, $y$ denotes the true label (real or fake), $\hat{y}$ represents the model's predicted probability of the news article belonging to the `fake' class. This loss function measures the model's prediction performance and guides the optimization process. Minimizing binary cross-entropy drives the model to refine its predictions, resulting in an enhanced ability to discriminate between fake and real news.
\section{Experimental Evaluations}
\label{sec:exp}
In this section, we elucidate the experimental design and subsequently provide a rigorous analysis of the results, demonstrating the efficacy of our proposed method.
\subsection{Experimental Setup}
This section commences with a detailed exposition of the various datasets employed for the experimental evaluation. Following this, we present a discourse on the baseline methods chosen for comparative analysis. Finally, the evaluation metrics utilized to assess the performance of our proposed method are delineated.
\subsubsection{Datasets}
\begin{table}[H]
\centering
\caption{Summary Statistics for Datasets}
\label{table:data_stats} 
\begin{tabular}{lcccccc}
\hline
\textbf{Dataset} & \textbf{Modality}&{\textbf{Languages}}&{\textbf{Real}} &{\textbf{Fake}}&{\textbf{Total}}\\
\midrule
MMIFND &Text+Image & 7 &14335 & 13750& 28085\\
Tamil & Text&1&2324 &2949 & 5313\\
\bottomrule
\end{tabular}
\end{table}
We employed two different datasets to validate the effectiveness of our approach. The statistics for different datasets are summarized in \autoref{table:data_stats}.
\paragraph{MMIFND} This dataset, a novel contribution, comprises news samples in seven distinct Indic languages: Hindi, Gujarati, Marathi, Bengali, Tamil, Punjabi, and Malayalam. The linguistic selection reflects the widespread geographic distribution of these languages within the Indian subcontinent. Our dataset incorporates a multimodal approach featuring both textual and visual components. MMIFND exhibits a near-balanced distribution, with 51.04\% real samples and 48.96\% fake samples.
\paragraph{Tamil Dataset} This dataset was originally developed by Mirnalinee \textit{et al.}~\cite{Mirnalinee2023}. News excerpts were extracted from various media channels and classified as authentic or fabricated. These news items were subsequently categorized through human annotation into a five-class taxonomy comprising Sports, Politics, Science, Entertainment, and Miscellaneous. The dataset exhibits a near-equitable distribution, with 55.9\% classified as fabricated and 44.1\% as authentic. Notably, this dataset is unimodal (textual) and exclusively focused on the Tamil language.
\subsubsection{Compared Methods}
To benchmark the performance of our proposed model, we compare its fake news detection capabilities with the following existing methods.
\begin{enumerate}
\item{SpotFake}~\cite{Singhal2019}: It adopts a multimodal approach to enhance fake news detection. It extracts textual features using a language model (BERT) and visual features utilizing Visual Geometry Group (VGG19) model~\cite{simonyan2014very}. These feature representations are then concatenated to form a unified vector for classification.
\item{Semi-FND}~\cite{Singh2023}: This work introduced a novel multimodal stacked ensemble-based approach (SEMIFND) for faster fake news detection with reduced parameters. Unimodal analysis revealed NASNet as the optimal image model. An ensemble of BERT and ELECTRA~\cite{clark2020electra} provides textual features. Decision-level fusion combines modalities and stacked text models. The proposed framework demonstrates reduced compute requirements and decision time while maintaining accuracy for the task of multimodal fake news detection.
\item{Mul-FAD}~\cite{Ahuja2023}: The authors developed a hierarchical attention-based network for multilingual fake news detection. They compiled a trilingual dataset (English, German, French) and employed cross-lingual fastText embeddings. Mul-FaD leverages a modified hierarchical attention architecture with TextCNN and stacked GRUs, enabling word-level and sentence-level analysis for nuanced contextual representations.
\item{HFND-TE}~\cite{Praseed2023}: This work introduced a novel ensemble-based approach for Hindi fake news detection. It leverages pre-trained transformer models (XLM-RoBERTa, mBERT, ELECTRA) that are individually fine-tuned for the task. A majority voting strategy aggregates their outputs for robust classification.
\item{FND-CLIP}~\cite{Zhou2023}: The authors  introduced a multimodal fake news detection network utilizing Contrastive Language-Image Pretraining (CLIP). It employs unimodal encoders for textual and visual feature extraction, along with pairwise CLIP encoders to capture intermodal relationships. CLIP-generated features are weighted by similarity scores, indicating the alignment between text and image. An attention mechanism adaptively integrates textual, visual, and fused features for optimal fake news classification.
\end{enumerate}
\subsubsection{Evaluation Metrics}
The comprehensive assessment of fake news detection models necessitates using metrics illuminating performance from various perspectives. The key terms are defined as follows.
\textit{True Positive (TP):} An instance of fabricated news correctly classified as fake.
\textit{True Negative (TN)}: An instance of authentic news correctly classified as real.
\textit{False Positive (FP):} An instance of authentic news erroneously classified as fake.
\textit{False Negative (FN):} An instance of fabricated news erroneously classified as real.
\begin{enumerate}
\item{Accuracy (A):} It represents the overall proportion of correct classifications (both true positives and true negatives) relative to the total number of news instances examined.
\begin{equation}
A=(TP + TN) / (TP + TN + FP + FN)
\label{acc}
\end{equation}
\item{Precision (P):} It indicates the proportion of correctly identified fake news instances among all instances the model classified as fake. It prioritizes minimizing false positives.
\begin{equation}
    P=TP / (TP + FP)
    \label{pre}
\end{equation}
\item{Recall (R):} It measures the proportion of correctly identified fake news instances relative to the total number of actual fake news instances. It prioritizes minimizing false negatives.
\begin{equation}
    R=TP / (TP + FN)
    \label{recall}
\end{equation}
\item{F1-Score (F1):} It represents the harmonic mean between precision and recall. F1-score offers a balanced metric for assessing a model's ability to discern fake news, especially when a compromise between minimizing false positives and false negatives is desired.
\begin{equation}
    F1=2*(P* R)/(P + R)
    \label{f1-score}
\end{equation}
\end{enumerate}
\subsection{Experimental Results}
This section delves into the experimental outcomes achieved by both existing and proposed methods. Here, we assess the efficacy of our proposed method by juxtaposing its performance against state-of-the-art methods on various datasets. Additionally, we conduct an ablation study to evaluate different feature combinations and the contributions of individual components. We conduct qualitative analysis and provide implementation details to offer deeper insights into our method.
\subsubsection{Effectiveness Comparison on MMIFND}
\begin{table}
\centering
\caption{Effectiveness Comparison Results on MMIFND}
\label{table:res_mmifnd} 
\resizebox{\columnwidth}{!}{ 
    \begin{tabular}{lccccccc}
\toprule
\textbf{Method} & \textbf{A} & \multicolumn{3}{c}{\textbf{Fake News}} & \multicolumn{3}{c}{\textbf{Real News}} \\ 
\cmidrule(r){3-5} \cmidrule(r){6-8} 
 &  & \textbf{P} & \textbf{R} & \textbf{F1} & \textbf{P} & \textbf{R} & \textbf{F1} \\ 
\midrule
SpotFake \cite{Singhal2019} & 0.972 & 0.957 & 0.989 & 0.973 & 0.989 & 0.956 & 0.972 \\
Semi-FND \cite{Singh2023} & 0.978 & 0.979 & 0.976 & 0.977 & 0.977 & 0.979 & 0.978 \\
Mul-FaD \cite{Ahuja2023} & 0.955 & 0.967 & 0.927 & 0.947 & 0.942 & 0.977 & 0.959 \\
HFND-TE \cite{Praseed2023}& 0.961 & 0.993 & 0.927 & 0.959 & 0.933 & 0.993 & 0.962 \\ 
FND-CLIP \cite{Zhou2023} & 0.988 & 0.992 & 0.983 & 0.987 & 0.984 & 0.992 & 0.988 \\
MMCFND & \textbf{0.996} & \textbf{0.998} & \textbf{0.994} & \textbf{0.996} & \textbf{0.995} & \textbf{0.998} & \textbf{0.997} \\
    \bottomrule     
    \end{tabular}  
    }
\end{table}
\begin{figure*}[!h]
\includegraphics[width=\textwidth, height=6cm]{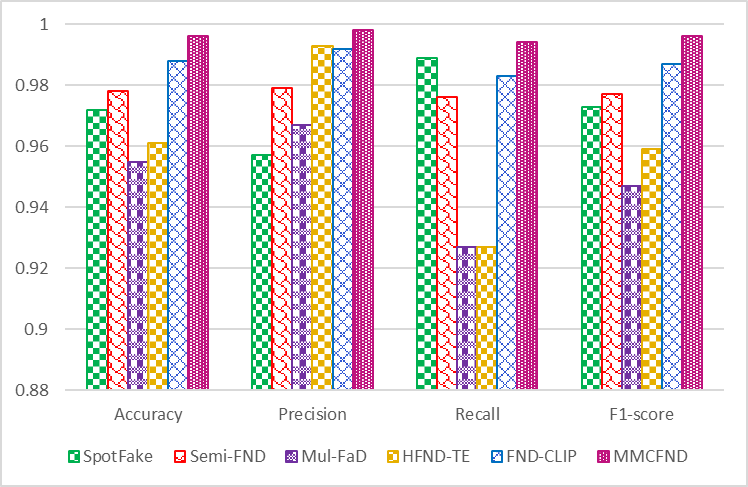}
	\caption{\centering Effectiveness Comparison of Different Methods on Fake Samples in MMIFND}
	\label{fig:mmcfnd_fake}
\end{figure*}
\begin{figure*}[!h]
\includegraphics[width=\textwidth,height=8cm]{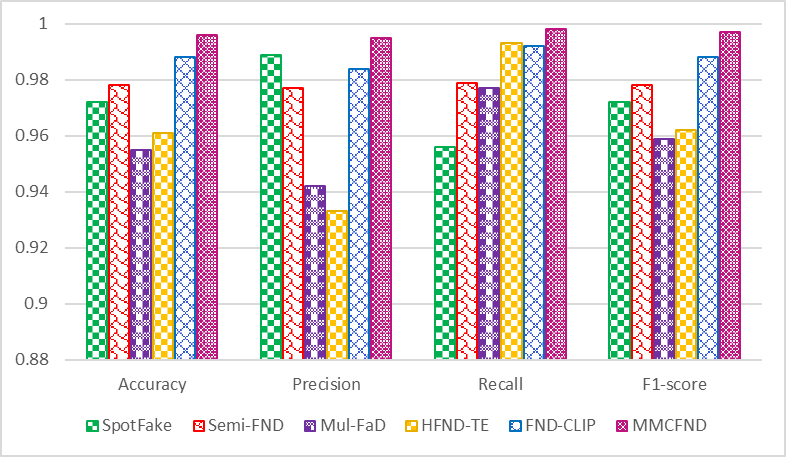}
	\caption{\centering Effectiveness Comparison of Different Methods on Real Samples in MMIFND}
	\label{fig:mmcfnd_real}
\end{figure*}
To evaluate the effectiveness of our proposed model in detecting fake news within low-resource Indic languages, we conducted a comprehensive comparative analysis with existing methods. Aligned with established research in this field, we employed various performance metrics such as accuracy, precision, recall, and F1-score, calculated for real and fake news samples in order to assess the efficacy of different approaches. 

\autoref{table:res_mmifnd} summarizes the results achieved by our model and existing methods on the MMIFND dataset. As evident from \autoref{table:res_mmifnd}, our model demonstrates a significant performance improvement compared to prior state-of-the-art approaches on the MMIFND dataset. MMCFND outperforms SpotFake, exhibiting gains of 2.4\% in accuracy, 2.3\% in F1-score (fake), and 2.5\% in F1-score (real). This improvement can be attributed to several factors. Firstly, MMCFND benefits from NASNet's superior image feature extraction capabilities. Secondly, MuRIL likely offers advantages in deriving textual features from multilingual news articles. Lastly, FLAVA provides a more sophisticated approach to multimodal feature fusion than SpotFake's simple concatenation, enabling MMCFND to develop a more comprehensive understanding of the interplay between text and image modalities. MMCFND achieves improvements of 1.8\% in accuracy, 1.9\% in F1-score (fake), and 1.9\% in F1-score (real) over Semi-FND. These performance gains are due to MMCFND leveraging Semi-FND's identification of NASNet as the optimal image model. Additionally, FLAVA provides MMCFND with a more nuanced understanding of multimodal relationships than Semi-FND's decision-level fusion. Furthermore, image captions generated by BLIP-2 provide MMCFND with valuable additional textual context. MMCFND demonstrates notable enhancements compared to Mul-FAD, with an increase of 4.1\% in accuracy, 4.9\% in F1-score (fake), and 3.8\% in F1-score (real). This is due to MMCFND's employment of advanced models like NASNet and MuRIL for superior visual and textual feature extraction, surpassing Mul-FAD's architecture based on TextCNNs and stacked Gated Recurrent Unit (GRU). MMCFND exhibits performance advantages over HFND-TE, with improvements of 3.5\% in accuracy, 4.9\% in F1-score (fake), and 3.8\% in F1-score (real). This can be attributed to MMCFND's multimodal approach, providing a richer representation than HFND-TE's text-only focus. Additionally, MMCFND's specific ensemble composition and decision-level fusion strategy is more optimized for this task. MMCFND surpasses FND-CLIP with improvements of 0.8\% in accuracy, 0.9\% in F1-score (fake), and 0.9\% in F1-score (real). Although FND-CLIP employs CLIP for multimodal interactions, MMCFND's use of FLAVA potentially offers a more sophisticated approach to feature fusion. FLAVA enables MMCFND to derive a richer multimodal representation, capturing subtle interactions between visual and textual elements that FND-CLIP's feature extraction might overlook. MMCFND's use of BLIP-2 for image caption generation likely plays a crucial role by providing additional textual context, leading to a more comprehensive understanding of the news article. \autoref{fig:mmcfnd_fake} and \autoref{fig:mmcfnd_real} depict the comparative performance of different fake news detection methods in relation to accuracy, precision, recall, and F1-score on both fake and real news samples on MMIFND dataset, respectively. Notably, the curves of the MMCFND model consistently surpass those of the existing methods across all metrics, highlighting the competitive advantage and efficacy of our proposed model.
\subsubsection{Effectiveness Comparison on  Tamil Dataset}
\begin{table}[H]
\centering
\caption{Effectiveness Comparison on Tamil Dataset}
\label{table:res_tamil} 
\begin{tabular}{lccccc}
\hline
\textbf{Method} & \textbf{A} & \textbf{P} & \textbf{R} & \textbf{F1}\\
\midrule
SpotFake \cite{Singhal2019} & 0.871 & 0.876 & 0.876 & 0.871 &  \\
Semi-FND \cite{Singh2023} & 0.950 & 0.949 & 0.950 & 0.950 &  \\
Mul-FaD \cite{Ahuja2023} & 0.930 & 0.929 & 0.929 & 0.929 &  \\
HFND-TE \cite{Praseed2023}& 0.955 & 0.954 & 0.957 & 0.954 &  \\
FND-CLIP \cite{Zhou2023} & 0.952 & 0.955 & 0.949 & 0.951 &  \\
\textbf{MMCFND} & \textbf{0.964} & \textbf{0.963} & \textbf{0.965} & \textbf{0.964} & \textbf{} \\
\bottomrule
\end{tabular}
\end{table}
To assess the efficacy of our proposed model in tackling the challenge of text-only fake news detection within low-resource Indic languages, we conducted a comparative evaluation using the text-only Tamil Dataset. 
\begin{figure*}
\includegraphics[width=\textwidth,height=8cm]{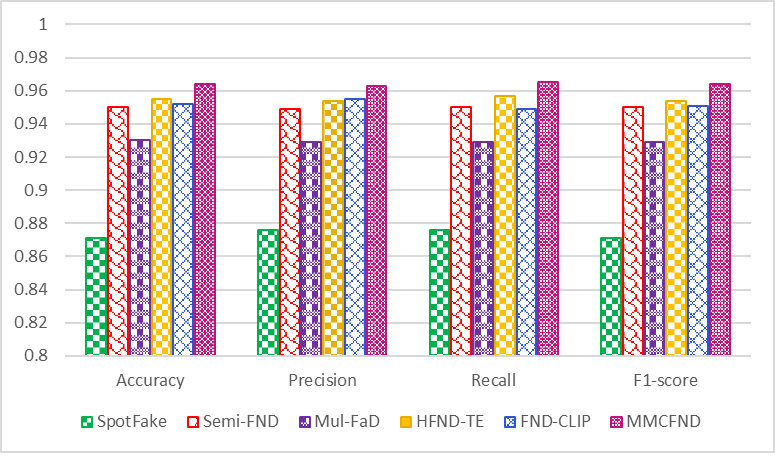}
	\caption{\centering Effectiveness Comparison of Different Methods on Tamil Dataset}
	\label{fig:tamil_sota}
\end{figure*}
\autoref{fig:tamil_sota} depicts that MMCFND surpasses existing methods when relying solely on textual information for fake news identification. From \autoref{table:res_tamil}, it can be seen that our proposed model achieves statistically significant improvements compared to existing methods on both accuracy and F1-score. MMCFND outperforms SpotFake by a substantial margin of 9.3\% in both accuracy and F1-score. Compared to Semi-FND, our model exhibits improvements of 1.4\% for both metrics. Similarly, significant advancements are observed against Mul-FaD (3.4\% and 3.5\%), HFND-TE (0.91\% and 1.0\%), and FND-CLIP (1.2\% and 1.3\%). These findings demonstrate the effectiveness of MMCFND in accurately identifying fake news written in low-resource Indic languages. MMCFND outperforms SpotFake due to MuRIL's specialized Tamil understanding, FLAVA's potential, and its ensemble approach, which provides a more nuanced analysis than BERT alone. While Semi-FND also uses an ensemble, MMCFND has an advantage since its ensemble composition i.e., MuRIL and FLAVA's text encoder is better suited for the Tamil dataset. MuRIL's specialized pre-training and FLAVA's potential benefit gives MMCFND an edge over Mul-FaD's TextCNNs and stacked GRUs in feature representations on the Tamil dataset. MMCFND's ensemble of diverse pre-trained models captures a wider range of linguistic patterns than HFND-TE's ensemble, which employs XLM-RoBERTa, mBERT, ELECTRA. The specialized Tamil pre-training of MuRIL and its combination with FLAVA within MMCFND's ensemble offers advantages over FND-CLIP's general-purpose text encoders.

\subsubsection{Ablation Studies}
This section elaborates on experiments meticulously designed to evaluate the relative significance of various modalities for detecting multimodal fake news articles written in low-resource Indic languages. 
\begin{table}
\centering
\caption{Modality Combinations on MMIFND}
\label{table:modality_combo}   
    \begin{tabular*}{\columnwidth}{@{\extracolsep{\fill}} lcccccc }
    \toprule
        \textbf{Methods} & \multicolumn{3}{c}{\textbf{Fake News}} & \multicolumn{3}{c}{\textbf{Real News}} \\ 
        \cmidrule(l){2-4} \cmidrule(l){5-7}
            &\textbf{P}&\textbf{R} &\textbf{F1}
            &\textbf{P}&\textbf{R} &\textbf{F1}\\  
    \midrule
        w/o Image & 0.986 & 0.941 & 0.963 & 0.945 &0.986& 0.965\\
        w/o Text &0.985 & 0.834 &0.903 & 0.858 & 0.987 & 0.918\\
        \textbf{Text+Image} & \textbf{0.998} & \textbf{0.994} & \textbf{0.996} & \textbf{0.995} & \textbf{0.998} &\textbf{0.997}\\
    \bottomrule     
    \end{tabular*}   
\end{table}
To systematically assess the contribution of each feature type, a rigorous ablation study was conducted. Our proposed model leverages a rich feature space encompassing images, texts, and captions to enhance the detection of fabricated news articles. The performance comparison of different modality combinations is shown in \autoref{table:modality_combo}. 
Removing textual features resulted in a substantial decline in the F1-score for fake (9.3\%) and real samples (7.9\%). This finding underscores the paramount role of linguistic cues in discerning deceptive content from factual information. Visual content also positively contributed to the model’s performance, albeit to a lesser extent. Excluding image features led to a moderate decrease in the F1-score for both fake (3.3\%) and real samples (3.2\%). The ``Text+Image" row has the highest F1-scores for both fake news and real news detection. This demonstrates that combining both image and text modalities leads to the most accurate fake news detection model.
\subsubsection{Effectiveness of Different Components}
This section elucidates the key components of our proposed model and their contributions to robust fake news detection. 
\begin{table}
\centering
\caption{Effectiveness of FLAVA-guided Feature Retrieval}
\label{table:flava_ablation}   
    \begin{tabular*}{\columnwidth}{@{\extracolsep{\fill}} lcccccc }
    \toprule
        \textbf{Methods} & \multicolumn{3}{c}{\textbf{Fake News}} & \multicolumn{3}{c}{\textbf{Real News}} \\ 
        \cmidrule(l){2-4} \cmidrule(l){5-7}
            &\textbf{P}&\textbf{R} &\textbf{F1}
            &\textbf{P}&\textbf{R} &\textbf{F1}\\        
    \midrule
        {w/o FLAVA} & 0.995 & 0.971 & 0.983 & 0.972 & 0.996 & 0.984\\
        \textbf{with FLAVA} & \textbf{0.998} & \textbf{0.994} & \textbf{0.996} & \textbf{0.995} & \textbf{0.998} &\textbf{0.997}\\
    \bottomrule     
    \end{tabular*}   
\end{table}
As seen from \autoref{table:flava_ablation}, the integration of multimodal features, meticulously extracted by FLAVA resulted in a marginal yet noteworthy improvement in F1-score compared to a model relying solely on unimodal textual and visual feature representations. The removal of these multimodal features leads to a decrease of 1.3\% for both fake and real samples. This finding suggests that multimodal features offer valuable supplemental cues for detecting deceptive content. FLAVA's ability to process and align textual and visual information provides a more holistic representation of the news article's content. This comprehensive approach mitigates the risk of manipulation through text alone, a crucial consideration in low-resource languages where linguistic cues might be less abundant or readily interpretable. By augmenting textual analysis with visual data, FLAVA constructs a representation that is more resilient to deceptive tactics. Furthermore, in low-resource languages where the vocabulary is more constrained, visuals serve to bridge semantic gaps. Concepts that may be difficult to express linguistically can be conveyed more effectively through imagery. Moreover, cross-cultural nuances like humor, metaphors, or references can be lost in translation when relying solely on textual analysis. FLAVA's multimodal approach mitigates this risk by providing a culturally-agnostic visual representation. 

Overall, FLAVA's integrated analysis of text and visuals demonstrates the potential for a more comprehensive and manipulation-resistant understanding of content. This approach holds particular promise for low-resource languages, where multimodal analysis can compensate for inherent linguistic limitations and provide a powerful tool for combating the spread of misinformation.
\begin{table}[H]
\centering
\caption{Effectiveness of BLIP-2 Component}
\label{table:blip_ablation} 
\begin{tabularx}{\columnwidth}{@{\extracolsep{\fill}} l *{6}{c} } 
    \toprule
        \textbf{Methods} & \multicolumn{3}{c}{\textbf{Fake News}} & \multicolumn{3}{c}{\textbf{Real News}} \\ 
        \cmidrule(l){2-4} \cmidrule(l){5-7}
            &\textbf{P}&\textbf{R} &\textbf{F1}
            &\textbf{P}&\textbf{R} &\textbf{F1}\\        
    \midrule
        {w/o BLIP-2}& 0.994 & 0.991 & 0.992 & 0.993 & 0.995 & 0.994\\
        \textbf{with BLIP-2} & \textbf{0.998} & \textbf{0.994} & \textbf{0.996} & \textbf{0.995} & \textbf{0.998} &\textbf{0.997}\\
    \bottomrule     
    \end{tabularx}   
\end{table}
The component analysis further delved into the caption generation component (BLIP-2) within MMCFND. The gain in performance obtained on inclusion of BLIP-2 is shown in \autoref{table:blip_ablation}. 
BLIP-2 distills the semantic relationships within the multimodal data, providing an additional layer of contextual information. This enhanced understanding is particularly important in nuanced cases where subtle inconsistencies between the visual and textual content may reveal the fabricated nature of the news. BLIP-2 strengthens the robustness of the MMCFND model for multimodal Indic fake news detection.
\subsubsection{Qualitative Analysis}
This section presents a qualitative analysis that evaluates the accuracy and strength of the predictions derived from our proposed model.
\begin{figure*}
\centering
\subfloat[Example 1]{%
\includegraphics[width=0.45\textwidth,height=17cm]{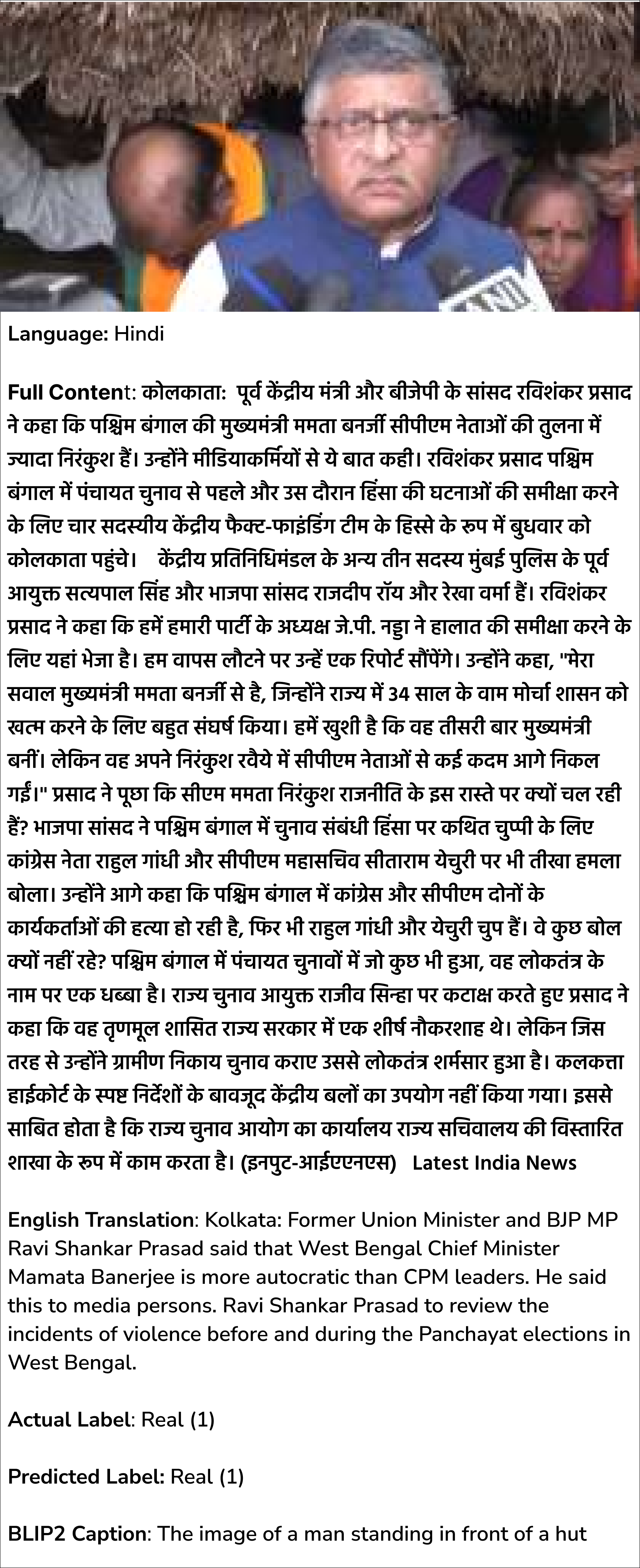}\label{fig:qa1}%
}
\subfloat[Example 2]{%
  \includegraphics[width=0.45\textwidth,height=17cm]{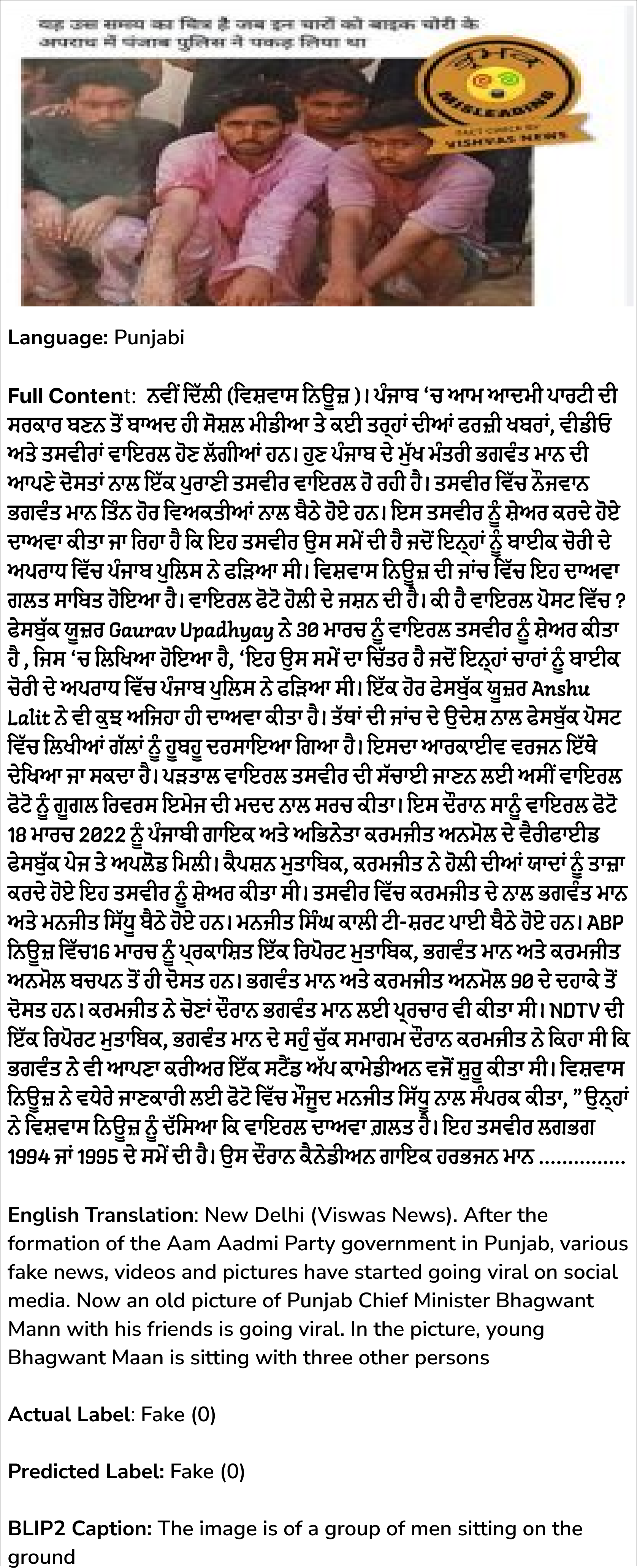}\label{fig:qa2}%
}
\caption{Fake News Samples Depicting Predictions of MMCFND}
\label{fig:qa}
\end{figure*}
As depicted in the example post~\autoref{fig:qa1}, MMCFND successfully predicts the correct label (``Real") for the given post. This can be attributed to the model's ability to effectively capture and integrate information from various modalities, such as text, image, and caption. The textual content of the post supports the assertion that Ravi Shankar Prasad is criticizing Mamata Banerjee for her autocratic rule. Phrases like ``more autocratic than CPM leaders", ``niramkush ravaiye" (autocratic attitude), and ``niramkush rajniti" (autocratic politics) are included. The image portrays Ravi Shankar Prasad standing in front of a hut. This backdrop represents the common people or rural areas of West Bengal, aligning with Prasad's critique of Banerjee's alleged mistreatment of these groups. Although the BLIP-2 generated caption, ``the image of a man standing in front of a hut," does not directly relate to specific political statements, it provides additional context that is subtly pertinent. The mention of a ``hut" reinforces the notion of rural Bengal and the common people, indirectly supporting Prasad's narrative. Therefore, the generated caption is relevant to the overall assessment of the post's authenticity. 
It accurately describes the primary visual elements of the image, aiding comprehension. Furthermore, it maintains a neutral tone, avoiding the introduction of biases or interpretations that could influence the detection of fake news. Therefore, the model's ability to effectively combine information from multiple modalities and the relevance of the generated caption contributes to its accurate prediction of the post's label as ``Real." In another exemplar post illustrated in \autoref{fig:qa2}, MMCFND correctly identifies the given Hindi news post as ``Fake" (0). The post falsely alleges that Punjab Chief Minister Bhagwant Mann was arrested for bike theft. The model's ability to process and integrate information from multiple sources, including text, image, and image caption, was pivotal in making this accurate prediction. The textual content of the post makes the false claim of Mann's arrest without providing any corroborating evidence. Furthermore, the mention of bike theft and the involvement of the Punjab Police further reinforced the fabricated narrative. The image depicts Mann sitting on the ground with friends, dressed in colorful Holi attire. This setting does not indicate the presence of police or an arrest. The BLIP-2 generated caption, ``the image is of a group of men sitting on the ground", accurately describes the visual content but does not directly address the fake news claim. Nevertheless, the model effectively integrates information from multiple modalities to ascertain the post's falsity. The text's false allegations, the image's lack of supporting evidence, and the neutral image caption all contribute to the accurate prediction.
\subsubsection{Implementation Details} The experiments were conducted on  a Linux server equipped with an Intel(R) Xeon(R) Silver 4215R CPU operating at 3.20GHz, 256 GB of RAM, and a 16 GB NVIDIA Tesla T4 GPU. Datasets were partitioned using an 80:20 train: test split. ReLU activation functions and a dropout probability of 0.2 were consistently applied within the model's fully connected layers. The model was trained with a batch size of 64, employing the Adam optimizer with a learning rate of 1e-3. Our novel dataset, MMIFND is available at \url{https://github.com/shubhi-bansal/MMCFND}.
\section{Conclusion}
\label{sec:con}
The scarcity of annotated data in low-resource Indic languages is a major obstacle in fighting multimodal fake news. To address this challenge, we introduce the first large-scale, \textbf{M}ultimodal \textbf{M}ultilingual dataset for \textbf{I}ndic \textbf{F}ake \textbf{N}ews \textbf{D}etection 
 (MMIFND) that covers seven low-resource languages. This dataset directly tackles the data scarcity issue, providing a valuable resource for developing and rigorously evaluating fake news detection methods tailored to these underrepresented languages. Moreover, we propose a novel \textbf{M}ultimodal \textbf{M}ultilingual \textbf{C}aption-aware framework designed for \textbf{F}ake \textbf{N}ews \textbf{D}etection (MMCFND) in Indic languages. MMCFND utilizes a vision and language foundation model to retrieve crossmodal image-text representations. It leverages powerful pre-trained encoders and a two-stage encoding process for in-depth analysis. Additionally, we generate descriptive image captions to enrich the feature space. The framework's four-pathway architecture extracts features from various modalities, which are then fed into a classifier to determine the veracity of news articles. Thorough experimentation on MMIFND demonstrates that MMCFND outperforms existing methods in extracting the most relevant features for accurate fake news detection in Indic languages. Our research offers significant advancements, with MMIFND as a novel dataset and MMCFND presenting a robust analytical framework. 

While MMCFND represents a significant advancement in fake news detection, there are still limitations that need to be addressed. The framework's reliance on pre-trained models may introduce potential biases inherited from their original training data, which could affect generalizability. Future research should focus on mitigating biases in pre-trained models and increasing dataset diversity. Expanding MMIFND's coverage to include a broader range of Indic languages would greatly enhance its applicability and impact. Integrating Optical Character Recognition (OCR) techniques would enable the framework to handle text embedded within images, thus broadening its usability in real-world scenarios.


 
%

\bibliographystyle{IEEEtran}
\bibliography{bibiliography}



 




\vfill

\end{document}